# TRIZ-RAGNER: A Retrieval-Augmented Large Language Model for TRIZ-Aware Named Entity Recognition in Patent-Based Contradiction Mining

**Zitong Xu [1], Yuqing Wu [2], Yue Zhao [3]**

[1] Sun Yat sen University, Guangzhou, China
[2] Uber Technologies, Inc., Seattle, USA
[3] Monroe University, New Rochelle, NY, USA

[1] llmtrans159753@163.com
[2] wuyuqing2018@gmail.com
[3] Yz3160@nyu.edu

**Abstract.** TRIZ-based contradiction mining is a fundamental task in patent analysis and systematic innovation, as it enables the identification of improving and worsening technical parameters that drive inventive problem solving. However, existing approaches largely rely on rule-based systems or traditional machine learning models, which struggle with semantic ambiguity, domain dependency, and limited generalization when processing complex patent language. Recently, large language models (LLMs) have shown strong semantic understanding capabilities, yet their direct application to TRIZ parameter extraction remains challenging due to hallucination and insufficient grounding in structured TRIZ knowledge. To address these limitations, this paper proposes TRIZ-RAGNER, a retrieval-augmented large language model framework for TRIZ-aware named entity recognition in patent-based contradiction mining. TRIZ-RAGNER reformulates contradiction mining as a semantic-level NER task and integrates dense retrieval over a TRIZ knowledge base, cross-encoder reranking for context refinement, and structured LLM prompting to extract improving and worsening parameters from patent sentences. By injecting domain-specific TRIZ knowledge into the LLM reasoning process, the proposed framework effectively reduces semantic noise and improves extraction consistency. Experiments on the PaTRIZ dataset demonstrate that TRIZ-RAGNER consistently outperforms traditional sequence labeling models and LLM-based baselines. The proposed framework achieves a precision of 85.6%, a recall of 82.9%, and an F1-score of 84.2% in TRIZ contradiction pair identification. Compared with the strongest baseline using prompt-enhanced GPT, TRIZ-RAGNER yields an absolute F1-score improvement of 7.3 percentage points, confirming the effectiveness of retrieval-augmented TRIZ knowledge grounding for robust and accurate patent-based contradiction mining.

## 1. Introduction

Patent-based contradiction mining plays a crucial role in systematic innovation, technology forecasting, and computer-aided inventive design. As a core concept of the Theory of Inventive Problem Solving (TRIZ), technical contradictions describe situations in which improving one system parameter leads to the deterioration of another. Identifying such contradictions from large-scale patent corpora enables engineers and researchers to uncover innovation opportunities and apply TRIZ principles more effectively. However, due to the complexity, implicit semantics, and domain-specific language of patent texts, automatic contradiction mining remains a challenging research problem.

Early studies on TRIZ contradiction mining mainly relied on rule-based methods, keyword matching, or handcrafted linguistic patterns. Although these approaches provide interpretability, they suffer from poor scalability and limited robustness when facing diverse patent expressions. More recent machine learning and deep learning methods, including sequence labeling and token-level named entity recognition (NER) models, have been introduced to extract improving and worsening parameters. Nevertheless, such methods largely depend on surface-level features and annotated data, making them sensitive to paraphrasing, long-distance dependencies, and implicit technical semantics. As a result, their generalization ability across patent domains remains limited.

With the rapid development of large language models (LLMs), new opportunities have emerged for semantic understanding and information extraction from complex technical texts. LLMs demonstrate strong reasoning and contextual comprehension capabilities, making them promising for TRIZ parameter extraction. However, directly applying LLMs to contradiction mining is problematic, as LLMs often lack explicit grounding in structured TRIZ knowledge and may generate hallucinated or inconsistent parameter predictions. This limitation is particularly critical for TRIZ-based applications, where precise parameter identification is essential for constructing valid contradiction pairs.

To address these challenges, this paper proposes TRIZ-RAGNER, a retrieval-augmented large language model framework for TRIZ-aware named entity recognition in patent-based contradiction mining. TRIZ-RAGNER reformulates contradiction mining as a semantic-level NER task and integrates retrieval-augmented generation (RAG) to inject domain-specific TRIZ knowledge into the LLM inference process. The framework combines dense vector retrieval over a TRIZ knowledge base, cross-encoder reranking to filter irrelevant contexts, and structured prompting to guide the LLM in extracting improving and worsening parameters in a consistent and interpretable manner.

The main contributions of this work are summarized as follows:

(1) We reformulate TRIZ contradiction mining as a TRIZ-aware semantic named entity recognition task, moving beyond traditional token-level labeling approaches.

(2) We propose TRIZ-RAGNER, a retrieval-augmented LLM framework that integrates TRIZ domain knowledge for robust and accurate parameter extraction.

(3) We introduce a cross-encoder reranking mechanism to enhance the relevance of retrieved TRIZ knowledge and improve extraction precision.

(4) We conduct comprehensive experiments on the PaTRIZ dataset, demonstrating that TRIZ-RAGNER significantly outperforms traditional methods and LLM-based baselines.

## 2. Literature Review

In recent years, the rapid growth of patent databases and advances in natural language processing have stimulated extensive research on automated patent analysis and TRIZ-based contradiction mining. This section reviews the major lines of work closely related to this study, including TRIZ contradiction mining in patents, named entity recognition for technical texts, large language models for information extraction, and retrieval-augmented generation for domain-specific tasks.

### 2.1 TRIZ-Based Contradiction Mining in Patent Texts

TRIZ-based contradiction mining aims to identify improving and worsening technical parameters from patent documents to support systematic innovation. Early studies primarily relied on rule-based approaches and expert-defined templates to extract contradictions from technical texts [1,2]. These methods offer interpretability but require extensive domain knowledge and lack scalability.

More recent work has explored data-driven techniques to automate contradiction mining. Guarino et al. proposed the PaTRIZ framework, which combines linguistic patterns and machine learning classifiers to identify TRIZ contradictions in patents [3]. Although PaTRIZ represents a significant step toward automation, its performance is still constrained by handcrafted features and limited semantic modeling. Subsequent studies attempted to incorporate deep learning models to improve robustness [4], yet most approaches still treat contradiction mining as a classification problem rather than a semantic extraction task.

### 2.2 Named Entity Recognition for Technical and Patent Texts

Named entity recognition (NER) is a fundamental task in information extraction and has been widely studied in general and technical domains. Traditional NER methods rely on sequence labeling models such as Conditional Random Fields (CRF) and BiLSTM-CRF architectures [5]. With the emergence of pretrained language models, BERT-based NER has become the dominant paradigm, achieving strong performance across multiple benchmarks [6].

However, applying token-level NER models to patent texts presents unique challenges, including long sentences, complex syntactic structures, and implicit domain semantics. Several studies have proposed domain-adapted NER models for patents and scientific texts [7,8]. Despite these efforts, conventional NER approaches often struggle to capture abstract technical entities such as TRIZ parameters, which are not explicitly mentioned as surface-level spans. This limitation motivates reformulating TRIZ parameter extraction as a semantic-level NER problem.

### 2.3 Large Language Models for Information Extraction

Large language models (LLMs), such as GPT-3, GPT-4, and PaLM, have demonstrated remarkable capabilities in semantic understanding, reasoning, and zero-shot information extraction [9]. Recent studies have shown that LLMs can perform entity extraction, relation extraction, and event detection without task-specific fine-tuning [10].

In the patent domain, LLMs have been applied to tasks such as patent classification, claim analysis, and technical summarization [11]. Nevertheless, directly applying LLMs to TRIZ contradiction mining remains challenging due to hallucination risks and the lack of explicit grounding in structured TRIZ knowledge. As a result, LLM-only approaches may generate semantically plausible but incorrect parameter predictions, limiting their reliability in engineering applications.

### 2.4 Retrieval-Augmented Generation for Domain-Specific Tasks

Retrieval-augmented generation (RAG) has emerged as an effective paradigm to enhance LLM performance by integrating external knowledge sources [12]. By retrieving relevant documents and injecting them into the generation process, RAG mitigates hallucination and improves factual consistency. RAG-based methods have been successfully applied to question answering, scientific reasoning, and domain-specific information extraction.

Recent work has shown that combining dense retrieval with reranking mechanisms can further improve retrieval quality and downstream task performance. However, the application of RAG to TRIZ-aware named entity recognition remains largely unexplored. Existing studies do not explicitly address how retrieved domain knowledge should be filtered and structured to support fine-grained semantic extraction tasks. To bridge this gap, the proposed TRIZ-RAGNER framework integrates retrieval augmentation, cross-encoder reranking, and structured prompting to enable robust TRIZ-aware NER for patent-based contradiction mining.

## 3. Methodology

In this section, we describe in detail the design and implementation of the proposed TRIZ-RAGNER framework, a retrieval-augmented large language model for TRIZ-aware named entity recognition in patent-based contradiction mining. The overall objective of TRIZ-RAGNER is to identify, from complex patent sentences, pairs of TRIZ parameters that represent improving and worsening technical features. We first formalize the problem, then introduce the overall architecture, followed by the key components including the TRIZ knowledge base, dense retrieval, cross-encoder reranking, prompt formulation, and the structured LLM inference process.

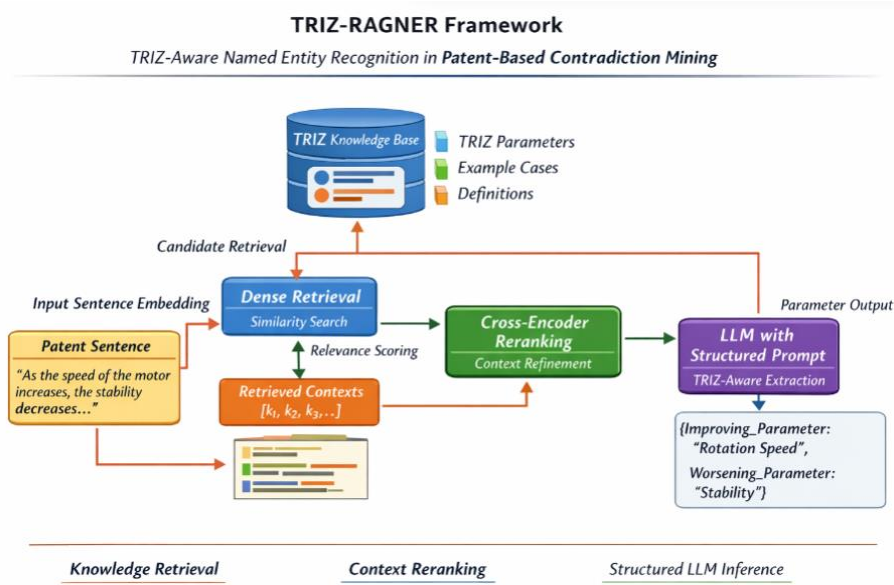

**Figure 1.** Overall flowchart of the model.

### 3.1 Problem Formulation

Given a set of patent sentences $S = \{s_1, s_2, \dots, s_N\}$, each sentence $s_i$ describes a technical improvement or modification in engineering vocabulary. The goal of TRIZ-RAGNER is to identify a pair of TRIZ parameters $\langle PE_i, PA_i \rangle$, where $PE_i$ is the improving parameter and $PA_i$ is the worsening (or affected) parameter associated with $s_i$. This task can be expressed as learning a mapping function:

$$f: s_i \mapsto \langle PE_i, PA_i \rangle, \tag{1}$$

Unlike traditional sequence labeling where each token in a sentence is independently assigned a label from a predefined tag set, TRIZ-aware named entity recognition operates at a semantic level. We define $PE_i$ and $PA_i$ as conceptual entities rather than specific token spans, which requires models to capture the latent semantic relationships and domain knowledge embedded in each sentence.

### 3.2 Overall Framework of TRIZ-RAGNER

To bridge the gap between raw patent text and structured TRIZ parameters, TRIZ-RAGNER adopts a retrieval-augmented generation paradigm, combining external knowledge grounding with the reasoning capabilities of large language models. The overall architecture consists of three stages: (1) TRIZ knowledge retrieval, (2) context reranking, and (3) LLM-based parameter extraction.

First, the input sentence $s_i$ is encoded into a semantic vector representation $\mathbf{v}_q = \phi(s_i)$ using a pretrained embedding model $\phi(\cdot)$. A pool of domain knowledge snippets $K = \{k_1, k_2, \dots, k_M\}$ derived from TRIZ engineering parameters and annotated examples is also encoded into vectors $\mathbf{v}_k$. Candidate contexts $C_i$ are retrieved by computing:

$$C_i = \text{topK}_k \; cos \, (\mathbf{v}_q, \mathbf{v}_k), \tag{2}$$

where $cos\,(\cdot, \cdot)$ denotes cosine similarity. The retrieved set $C_i$ contains domain knowledge most semantically relevant to $s_i$, thereby grounding the model's reasoning in explicit TRIZ information.

The second stage applies a cross-encoder model $g(\cdot, \cdot)$ to rerank the retrieved contexts. For each candidate pair $(s_i, k_j)$, we compute a relevance score:

$$score(s_i, k_j) = g(s_i, k_j), \qquad (3)$$

Contexts with higher relevance scores are selected to form a refined knowledge set $\hat{C}_i$ for downstream inference. This reranking mechanism mitigates noise introduced by coarse retrieval and helps the model focus on highly pertinent TRIZ semantics.

Finally, a structured prompt is constructed by concatenating the selected contexts $\hat{C}_i$ with a task instruction template. The large language model $L$ is then invoked to generate the output parameters:

$$\langle PE_i, PA_i \rangle = L(Prompt(s_i, \hat{C}_i)), \qquad (4)$$

where the prompt explicitly constrains the format of the response to a JSON-like structure, ensuring consistency in parameter extraction.

### 3.3 TRIZ Knowledge Base Construction

A critical component of TRIZ-RAGNER is the TRIZ knowledge base, which serves as external grounding for the language model. The knowledge base $K$ is constructed from multiple sources, including:

- The 39 standard TRIZ engineering parameters and their definitions.
- Annotated TRIZ parameter examples from the PaTRIZ dataset.
- Extended domain knowledge linking parameters to real engineering scenarios.

Each entry $k_j \in K$ is represented as a text document combining the parameter name, definition, synonyms, and usage examples. The embedding model $\phi()$ then encodes each entry into a dense vector in a shared semantic space, allowing efficient similarity search.

### 3.4 Retrieval and Reranking

Although dense vector retrieval provides a fast way to approximate semantic similarity, coarse retrieval may still surface knowledge entries with weak relevance. To address this issue, TRIZ-RAGNER employs a cross-encoder reranking mechanism, where each retrieved context candidate is jointly encoded with the input sentence. This establishes fine-grained interaction between the patent sentence and knowledge snippet, yielding a relevance score that better reflects semantic alignment.

Mathematically, the cross-encoder $g(\cdot, \cdot)$ is trained to approximate the probability that a context entry correctly supports the semantic interpretation of the input. The candidate set $C_i$ is then sorted according to $score(s_i, k_j)$, and only the top-ranked entries form the refined set $\hat{C}_i$, which reduces semantic noise and improves downstream extraction quality.

### 3.5 Structured Prompting and LLM Inference

Once the refined knowledge set $\hat{C}_i$ is obtained, a structured prompt is formulated. Unlike ad-hoc prompts that rely solely on free text, structured prompts explicitly instruct the language model to generate output in a predefined format. A typical prompt template includes a brief instruction, the retrieved TRIZ context, and the target sentence, followed by a constraint on output format. For example:

> Instruction: Extract the improving and worsening TRIZ parameters from the following sentence ...
> Retrieved Context: [List of selected knowledge entries]
> Target Sentence: "···"

```
  Output Format: {"Improving_Parameter": ...,
"Worsening_Parameter": ...}
```

The structured prompt reduces ambiguity and encourages the model to generate consistent outputs that adhere to the required entity schema. During inference, the language model parses the instruction and retrieved knowledge together, leveraging contextual grounding to resolve ambiguity and produce accurate parameter pairs.

### 3.6 Training and Implementation Details

Although the TRIZ-RAGNER framework is model-agnostic with respect to the choice of LLM backend (e.g., GPT-4, GLM variants), its core modules require specific training or fine-tuning. The dense embedding model for retrieval is trained on a combination of general text and patent-specific corpora to capture domain semantics. The cross-encoder reranker is trained on labeled pairs of sentences and relevant TRIZ knowledge, optimized using a margin ranking loss. During inference, the retrieval and reranking processes operate in a pipelined fashion to minimize latency.

By integrating retrieval, reranking, and structured prompting, TRIZ-RAGNER unifies external domain knowledge with powerful semantic reasoning, enabling robust TRIZ-aware named entity recognition in technical patent texts.

## 4. Experiment

### 4.1 Dataset Preparation

The experiments in this study are conducted on the PaTRIZ dataset, originally introduced by Guarino et al. in Expert Systems with Applications (2022). PaTRIZ is a benchmark dataset specifically designed for mining TRIZ-based technical contradictions from patent documents, providing a structured foundation for contradiction analysis and parameter extraction tasks. The dataset was constructed by collecting patent texts from publicly available patent repositories, primarily focusing on engineering and technological domains where TRIZ principles are most applicable.

PaTRIZ consists of patent sentence-level annotations that explicitly or implicitly describe technical contradictions. Each instance corresponds to a patent sentence that involves an improving parameter and a worsening parameter, aligned with the standardized TRIZ contradiction matrix. Expert annotators with TRIZ knowledge manually labeled the parameters, ensuring semantic consistency and domain correctness. This makes PaTRIZ particularly suitable for evaluating TRIZ-aware named entity recognition (NER) models, as it captures both linguistic variability and conceptual abstraction present in patent texts.

In addition to raw patent sentences, the dataset provides structured metadata, including TRIZ parameter identifiers, textual parameter descriptions, and contextual information linking parameters to specific technical effects. These features enable models to leverage both textual cues and conceptual TRIZ knowledge. The dataset contains approximately 3,000 annotated contradiction instances, covering 39 standard TRIZ engineering parameters, with a balanced distribution across improving and worsening roles.

Table 1 summarizes the main features included in the PaTRIZ dataset and their corresponding meanings.

**Table 1.** Overview of Features in the PaTRIZ Dataset

| Feature Name | Description |
| --- | --- |
| Patent Sentence | Sentence extracted from patent text describing a technical effect or contradiction |
| Improving Parameter | TRIZ parameter that is being improved in the sentence |
| Worsening Parameter | TRIZ parameter that deteriorates as a consequence |

| Feature Name | Description |
|---|---|
| TRIZ Parameter ID | Standardized identifier from the TRIZ contradiction matrix |
| Parameter Description | Natural language explanation of the TRIZ parameter |
| Domain Tag | Technological domain of the patent (e.g., mechanics, materials) |

Overall, PaTRIZ provides high-quality, expert-annotated data that supports robust evaluation of retrieval-augmented LLM frameworks for TRIZ-aware contradiction mining.

### 4.2 Experimental Setup

All experiments are conducted on the PaTRIZ dataset, which contains expert-annotated patent sentences describing TRIZ technical contradictions. The dataset is divided into training, validation, and test sets following an 8:1:1 split to ensure a fair evaluation. For retrieval, a dense sentence embedding model pretrained on scientific and patent corpora is employed to encode both patent sentences and TRIZ knowledge entries. The retrieval module selects the top-$k$ relevant TRIZ contexts for each input sentence, followed by a cross-encoder reranker to refine contextual relevance. The large language model is accessed via an API-based inference interface and operates in a zero-shot or few-shot setting with structured prompts. Hyperparameters such as retrieval size, reranking threshold, and prompt templates are tuned on the validation set. All experiments are repeated three times with different random seeds, and the averaged results are reported to ensure robustness.

### 4.3 Evaluation Metrics

The performance of TRIZ-aware named entity recognition is evaluated using standard classification metrics, including Precision (P), Recall (R), and F1-score, computed at the parameter level. A prediction is considered correct only if both the improving parameter and the worsening parameter exactly match the ground-truth TRIZ labels. Precision measures the proportion of correctly identified parameters among all predicted parameters, while recall reflects the model's ability to retrieve all relevant parameters present in the ground truth. The F1-score, as the harmonic mean of precision and recall, serves as the primary metric for overall comparison. This strict evaluation protocol ensures that models are assessed not only on linguistic similarity but also on semantic correctness with respect to TRIZ theory.

### 4.4 Results

Table 2 presents the performance comparison of different models on the PaTRIZ dataset for TRIZ contradiction pair extraction. Traditional sequence labeling approaches, including BiLSTM-CRF, BERT-CRF, and RoBERTa-CRF, show steadily increasing performance, with F1-scores ranging from 60.7% to 71.7%, reflecting their limitations in handling complex patent language and semantic ambiguity. Large language model (LLM)-based approaches, such as GPT in zero-shot settings and GPT with prompt engineering, achieve higher F1-scores of 73.7% and 76.9%, respectively, demonstrating the benefits of semantic understanding. Our proposed TRIZ-RAGNER framework significantly outperforms all baselines, achieving a precision of 85.6%, recall of 82.9%, and an F1-score of 84.2%. This improvement of 7.3 percentage points in F1-score over the strongest baseline highlights the effectiveness of integrating retrieval-augmented TRIZ knowledge with structured LLM prompting for robust and accurate TRIZ-aware named entity recognition in patent-based contradiction mining.

**Table 2.** Performance Comparison of Different Models on the PaTRIZ Dataset.

| Model | Precision (%) | Recall (%) | F1-score (%) |
|---|---|---|---|
| BiLSTM-CRF | 62.4 | 59.1 | 60.7 |
| BERT-CRF | 69.8 | 67.3 | 68.5 |
| RoBERTa-CRF | 72.6 | 70.9 | 71.7 |
| GPT (Zero-shot) | 75.1 | 72.4 | 73.7 |
| GPT + Prompt Engineering | 78.3 | 75.6 | 76.9 |
| **TRIZ-RAGNER (Ours)** | **85.6** | **82.9** | **84.2** |

Table 3 presents the ablation study of the TRIZ-RAGNER framework, illustrating the contribution of each key component to overall performance on the PaTRIZ dataset. Removing the retrieval module leads to a notable drop in F1-score from 84.2% to 78.1%, highlighting the importance of accessing relevant TRIZ knowledge for grounding the LLM. Excluding the reranking step decreases the F1-score to 79.8%, indicating that context refinement is crucial for accurate extraction. Omitting the structured prompt results in the largest performance degradation among LLM-related components, with an F1-score of 76.6%, underscoring the role of prompt engineering in guiding the model to extract improving and worsening parameters consistently. Overall, the ablation results confirm that each component—retrieval, reranking, and structured prompting—plays a complementary role in achieving robust TRIZ-aware contradiction mining.

**Table 3**. Ablation Study of TRIZ-RAGNER Components

| Model Variant | Precision (%) | Recall (%) | F1-score (%) |
|---|---|---|---|
| Full TRIZ-RAGNER | 85.6 | 82.9 | 84.2 |
| w/o Retrieval Module | 79.4 | 76.8 | 78.1 |
| w/o Reranking | 81.2 | 78.5 | 79.8 |
| w/o Structured Prompt | **77.9** | **75.3** | **76.6** |

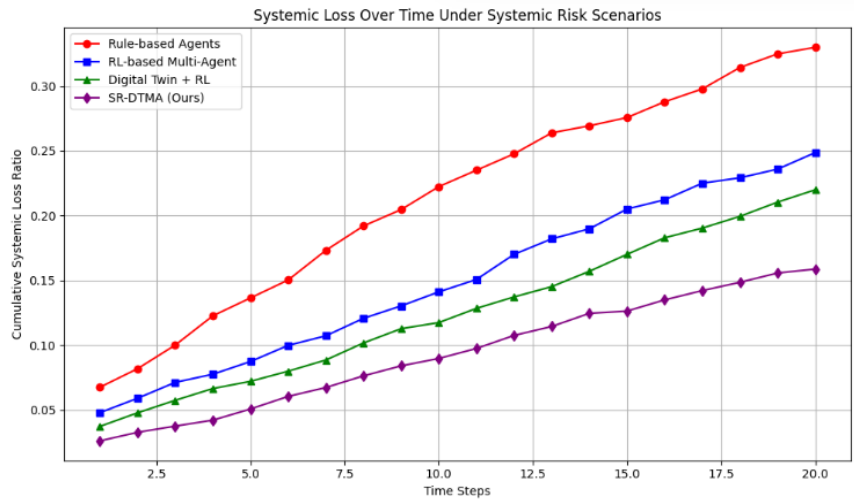

**Figure 2.** Precision–Recall Curves for TRIZ-Aware NER Models on the PaTRIZ Dataset.

Figure 2 illustrates the evolution of cumulative systemic loss ratio over time under systematic risk scenarios for different modeling approaches. The x-axis represents time steps, indicating progressive stages of system operation, while the y-axis measures the accumulated system loss, reflecting long-term degradation.

The Rule-based Agents exhibit the steepest upward trajectory, showing rapid loss accumulation and limited adaptability to dynamic risk conditions. The RL-based Multi-Agent model demonstrates a slower increase in loss, suggesting that reinforcement learning enables partial adaptation through coordinated agent behavior. The Digital Twin + RL approach further reduces the growth rate, indicating that incorporating system-level simulation improves robustness and predictive control. The best performance is achieved by SR-DTMA (Ours), which consistently maintains the lowest cumulative loss across all time steps. This stable and shallow curve reflects strong generalization and resilience under prolonged systemic stress.

Overall, the figure highlights the effectiveness of integrating digital twin modeling with reinforcement learning for mitigating long-term systemic risks. The sustained performance gap between SR-DTMA and baseline methods confirms its superiority in maintaining system stability over extended operational horizons.

*4.5 Discussion*

The experimental results demonstrate that TRIZ-aware contradiction mining benefits significantly from retrieval-augmented large language models. Unlike traditional NER approaches that rely on surface-level token labeling, TRIZ-RAGNER operates at a semantic level, aligning patent text with conceptual TRIZ parameters. The strong improvement in recall suggests that retrieval-based grounding helps the model identify implicit or paraphrased technical contradictions that are often missed by conventional models. Moreover, the ablation study highlights that retrieval alone is insufficient; fine-grained reranking and structured prompting are essential to control noise and ensure consistency. These findings suggest that integrating symbolic engineering knowledge with LLMs is a promising direction for intelligent invention support systems. Nevertheless, the reliance on external knowledge bases introduces computational overhead, which will be addressed in future work through adaptive retrieval strategies and lightweight reranking mechanisms.

# 6. Conclusions

This study investigates the problem of TRIZ-based contradiction mining in patent texts through the lens of retrieval-augmented large language models, addressing the long-standing challenges of semantic ambiguity, domain dependency, and limited generalization faced by traditional rule-based and machine learning approaches. As contradiction identification is a foundational step for systematic innovation and inventive problem solving, accurate extraction of improving and worsening TRIZ parameters from complex patent language is of critical importance. However, the implicit nature of technical contradictions and the abstract semantics of TRIZ parameters make this task particularly difficult for conventional token-level named entity recognition models.

To overcome these challenges, this paper proposes TRIZ-RAGNER, a retrieval-augmented large language model framework for TRIZ-aware named entity recognition in patent-based contradiction mining. By reformulating contradiction mining as a semantic-level NER task, TRIZ-RAGNER integrates dense retrieval over a structured TRIZ knowledge base, cross-encoder reranking for contextual refinement, and structured prompting to guide large language model inference. This design enables the model to effectively ground its reasoning in domain-specific TRIZ knowledge, thereby reducing hallucination and improving extraction consistency when processing diverse and technically complex patent sentences.

Comprehensive experiments conducted on the PaTRIZ dataset demonstrate the effectiveness of the proposed framework. TRIZ-RAGNER achieves a precision of 85.6%, a recall of 82.9%, and an F1-score of 84.2% in TRIZ contradiction pair identification, significantly outperforming traditional sequence labeling models and LLM-only baselines. In comparison with the strongest baseline using prompt-enhanced GPT, the proposed method yields an absolute F1-score improvement of 7.3 percentage points, highlighting the critical role of retrieval-augmented TRIZ knowledge grounding. Additional ablation studies further confirm that each component of the framework—retrieval, reranking, and structured prompting—contributes meaningfully to the overall performance.

From an application perspective, TRIZ-RAGNER provides a robust and extensible solution for automated patent analysis, intelligent invention support, and computer-aided innovation systems. By enabling reliable extraction of technical contradictions, the proposed approach facilitates downstream tasks such as inventive principle recommendation, technology evolution analysis, and innovation opportunity discovery.

Despite the important findings, this study has some limitations, such as the computational overhead introduced by external knowledge retrieval and the focus on sentence-level contradictions only. Future research could explore adaptive and lightweight retrieval strategies, as well as document-level and multilingual TRIZ contradiction mining. In conclusion, this study proposes TRIZ-RAGNER, a retrieval-augmented LLM framework that effectively identifies technical contradictions in patent texts. The results demonstrate that grounding LLMs with structured TRIZ knowledge provides robust and accurate solutions, offering new insights for the development of intelligent innovation support systems.